\documentclass{article}  
\title{
Selective State-Space Neural Tucker Factorization for Temporal-aware Traffic Imputation
}

\author{Zhenhao Li and Peng Tang}

\usepackage{amssymb}
\usepackage{cite}
\usepackage{makecell} 
\usepackage{amsmath}
\usepackage{booktabs}    
\usepackage{multirow}    
\usepackage{graphicx}     

\begin{document}

\maketitle
\thispagestyle{empty}
\pagestyle{empty}

\begin{abstract}
Spatiotemporal traffic data imputation is a key task of intelligent transportation systems (ITS), and urban traffic data inherently exhibits strong long-term temporal dependencies. Therefore, accurate modeling of these long-term temporal dynamics is the core prerequisite for high fidelity traffic data imputation. However, existing traffic data imputation methods still suffer from inherent limitations in long-sequence modeling, such as gradient vanishing, low efficiency, and insufficient capability to capture long-range temporal correlations. To address these issues, this paper proposes a Selective State Space Neural Tucker Factorization model (S3NTF) for robust traffic data imputation. This model combines the Selective State Space mechanism with Neural Tucker Factorization to capture complex spatial correlations and long-term temporal dynamics. A state space-based time encoder is proposed to effectively extracts long sequence dependencies with linear complexity, while the latent spatiotemporal feature interaction learning module models high-order cross modal interactions based on Tucker decomposition. Extensive experiments on six real traffic datasets have shown that our model outperforms existing models in terms of accuracy.
\end{abstract}

\section{INTRODUCTION}

Traffic observations provide a foundation for intelligent transportation systems, including traffic state estimation and forecasting. Surveys of traffic imputation and spatiotemporal pattern discovery describe the importance of recovering incomplete observations and exploiting spatial and temporal structure~\cite{zhang2024imputationsurvey,r100}. Traffic records collected by sensors and GPS devices can be organized as multidimensional arrays, but sensor failures and communication interruptions leave missing entries. Reliable imputation is therefore an important prerequisite for subsequent traffic analysis.

Low-rank tensor completion offers a structured approach to this problem. Autoregressive tensor completion, temporally constrained dimension-preserved completion, and temporally regularized latent factor analysis explicitly incorporate temporal information into traffic reconstruction~\cite{chen2022latc,r22,r83}. More generally, latent tensor factorization represents incomplete multidimensional observations through interactions among compact factors~\cite{r23,r33}. Biases, nonnegativity constraints, and regularization extend this framework~\cite{r9,r11,r91}, while neural factorization provides additional choices for learning latent interactions~\cite{r15,r16}.

Neural Tucker factorization (NeuTucF) expresses Tucker-style latent interactions within a neural architecture~\cite{r73}. Auto-encoding and self-attending extensions further investigate nonlinear interaction learning~\cite{r28,r104,r105}. To incorporate temporal dependencies, the gated-recurrent neural Tucker factorization model (GARNT), introduced in \emph{Temporal-Aware Neural Tucker Factorization for Traffic Data Imputation}, combines recurrent temporal encoding with the Tucker interaction framework~\cite{hou2025tntf}. This motivates investigating alternative temporal encoders while retaining the same underlying factorization structure.

Mamba introduces input-dependent selection into state space sequence modeling and supports a hardware-aware parallel scan~\cite{gu2024mamba}. Its use in traffic forecasting has been explored by DST-Mamba, SpoT-Mamba, and ST-Mamba~\cite{he2025dstmamba,choi2024spotmamba,shao2024stmamba}. These studies motivate examining selective temporal encoding for missing-value reconstruction within a neural tensor factorization model. Here, the encoder processes a sequence of learned temporal index embeddings, and the resulting representation participates in the reconstruction of individual tensor entries.

This paper proposes a selective state space neural Tucker factorization model (S3NTF) for traffic data imputation. Its main contributions are: 1) a Mamba-based temporal encoder that aggregates historical temporal embeddings through selective state space modeling; 2) an interaction module that combines the encoded temporal representation with the remaining mode embeddings through a Tucker-style outer product; and 3) an evaluation on six traffic datasets, together with an analysis of the temporal window and state dimension.

\section{RELATED WORK}

\subsection{Traffic Imputation and Related Spatiotemporal Tasks}
Traffic imputation research spans low-rank reconstruction, graph-based learning, and generative modeling. Bayesian temporal factorization offers a probabilistic factorization approach to multidimensional time series~\cite{chen2022btmf}, while low-rank autoregressive tensor completion and Hessian-regularized spatiotemporal low-rank modeling address traffic imputation~\cite{r71}. Laplacian convolutional representations, convolutional low-rank tensor representations for structural missingness, and low-rank models with deep plug-and-play priors provide further reconstruction strategies~\cite{chen2024lcr,li2024clrtr,chen2024pnp}. Robust low-rank latent feature analysis also studies spatiotemporal signal recovery~\cite{r40}.

Graph-based imputation is represented by GRIN and graph-aggregator-based traffic imputation~\cite{cini2022grin,liang2022tgan}. PriSTI investigates conditional diffusion for spatiotemporal imputation, whereas ImputeFormer combines low-rank structure with Transformer modeling~\cite{liu2023pristi,nie2024imputeformer}. A related but distinct setting is spatiotemporal kriging: INCREASE and KITS investigate inductive graph-based approaches, and Laplacian-enhanced low-rank tensor kriging studies sparse traffic sensing~\cite{zheng2023increase,xu2025kits,nie2023correlating}. Physics-informed traffic state estimation provides another perspective on incorporating domain structure~\cite{di2023pidlsurvey}.

Several neighboring tasks provide useful context for temporal representation learning. Hankel temporal matrix factorization and nonstationary temporal matrix factorization address traffic forecasting with sparse observations~\cite{chen2024htmf,chen2025ntmf}. Continuous-time graph representation learning and hierarchical-memory-based dynamic graph learning have been studied for origin--destination demand prediction~\cite{han2022cod,zhang2022dghm}. Tensor-based methods also appear in traffic anomaly detection and diagnosis~\cite{wang2022hrpca,zhao2024anomaly}. These works concern different inference targets and are discussed as related applications rather than as directly comparable imputation baselines.

\subsection{Latent Factorization of Incomplete Data}
Latent factorization has been investigated for large-scale dynamic weighted directed networks and hidden-pattern discovery~\cite{r14,r18}, as well as sparse recommendation matrices and biased nonnegative representations of incomplete data~\cite{r61,r50}. Within tensor models, instance-frequency-weighted regularization, temporal biases, and adaptively extended biases offer ways to represent incomplete dynamic QoS observations~\cite{r17,r13,r38}. Biased nonnegative Tucker factorization, learning-depth adjustment, and double-space/double-norm ensembles provide related modeling alternatives~\cite{r10,r21,r47}. These application-specific extensions supply background on factor design; they do not by themselves establish accuracy on traffic data.

Temporal factorization research includes temporally regularized sensing-network representations and temporal-relations-aware dynamic graph models~\cite{r94,r101}. Kalman-filter-based latent factor analysis and extended-Kalman-filter-based temporal QoS estimation investigate state estimation within latent representations~\cite{r42,r103}. Biased block-term decomposition and adaptive temporal-dependent low-rank modeling explore other forms of temporal structure~\cite{b3,b4}. These approaches complement the traffic-specific temporal constraints reviewed above.

Alternative tensor structures include biased nonnegative tensor-ring factorization and tensor-ring link prediction~\cite{r92,b5}, tensor-wheel representations, and extended CP tensor networks~\cite{b6,b9}. Efficient dynamic-network representations and proximal-ADMM-based factorization have also been studied~\cite{r32,r6}. Federated tensor factorization and federated latent factor learning extend the setting to privacy-preserving dynamic-network representation and spatiotemporal signal recovery~\cite{r107,r108}. S3NTF focuses on temporal encoding within a centralized neural Tucker model; these structural and deployment extensions are background rather than components of the proposed method.

\subsection{Neural and Graph-Enhanced Factorization}
Neural nonnegative tensor factorization, nonlinear CP factorization, and neural latent factorization for dynamic networks investigate neural representations of incomplete tensors~\cite{r1}. Neural-network-incorporated latent factor analysis and nonnegative autoencoders address related incomplete-data settings~\cite{r29,r44}. Multimetric autoencoders, prediction-sampling-based multilayer factor models, and multilayer randomized factor models offer further representation designs~\cite{r116,r48,r84}. Mode-aware Tucker networks and dynamic graph mixers extend the discussion to nonstandard tensors~\cite{r4,r7}.

NeuTucF supplies the Tucker-style latent interaction backbone used in this work. Auto-encoding neural Tucker factorization, attention-based neural tensor factorization, and multi-aspect or multi-projection self-attending Tucker models investigate richer interaction mappings~\cite{r45}. Temporal convolution is another relevant direction, including tensor causal convolution and dimension-preserved traffic prediction models~\cite{r8,r26}. Convolutional low-rank methods have also been explored for water-quality imputation~\cite{b10,b15}; their relevance here is methodological, since their application data differ from traffic observations.

Graph information can be combined with factorization through graph-incorporated or two-stream graph-convolutional latent feature analysis~\cite{r64,r53}. Spatiotemporal graph neural networks have likewise been incorporated into tensor factorization for dynamic QoS estimation~\cite{r96}. Convolution-bias-based tensor representations, global-dependency learning on dynamic graphs, and graph tensor convolution provide additional examples of structural modeling~\cite{r5,r25,r118}. This literature motivates distinguishing explicit graph propagation from the mode-wise latent interactions used by S3NTF.

\subsection{Optimization of Latent Factor Models}
Optimization is a complementary issue in learning from sparse observations. Related latent factor models investigate proximal and adaptive alternating-direction methods~\cite{r41,r55}, adaptive divergences~\cite{r72,r76}, and multimetric or mixed-norm objectives~\cite{r75,r89}. These works address the fitting procedure or objective rather than the selective temporal encoder studied here.

Acceleration strategies include generalized Nesterov methods for tensor and matrix factorization~\cite{r20,r80,r81}, Hessian-vector and accelerated second-order methods~\cite{r52,r77}, and distributed second-order factor learning~\cite{r57}. Controller-inspired methods include proportional-integral latent factor analysis~\cite{r34}, PID-incorporated tensor factorization~\cite{r12,r19}, fuzzy and nonlinear PID-guided stochastic gradient methods~\cite{r54,r79}, and adaptive PID-incorporated nonnegative factor analysis~\cite{r109}. Particle-swarm-based alternatives investigate pseudo-gradient adjustment, position transitions, and hierarchical search~\cite{r43,r46,r90}.

Parallel and asynchronous factor learning has been explored through adaptive parallel SGD~\cite{r74,r59}, asynchronous fuzzy SGD~\cite{r58}, multi-stream asynchronous SGD~\cite{r98}, and accelerated asynchronous parallel SGD~\cite{b12}. These studies provide context for the computational aspects of incomplete-data factorization. The experiments in this paper use Adam; the above optimizers are not implemented or compared here.

\subsection{Temporal Encoding and Selective State Spaces}
Spatiotemporal graph forecasting provides a broader context for temporal modeling~\cite{jin2024stgnnsurvey}. Examples include decoupled dynamic graph forecasting, propagation-delay-aware Transformers, and adaptive-embedding Transformers~\cite{shao2022d2stgnn,jiang2023pdformer,liu2023staeformer}. BigST investigates linear-complexity graph forecasting on large road networks~\cite{han2024bigst}. Within neural Tucker factorization, GARNT is the most direct temporal-encoding reference for the present study.

Mamba makes the SSM parameters $B$, $C$, and $\Delta$ input-dependent while retaining a learned, input-independent continuous-time matrix $A$; the discretized transition consequently depends on the input through $\Delta$. DST-Mamba, SpoT-Mamba, and ST-Mamba apply selective state space modeling to traffic or spatiotemporal graph forecasting. Their forecasting setting differs from the entry-wise reconstruction considered here. S3NTF studies selective encoding of temporal index embeddings within the Tucker interaction framework, with reconstruction loss evaluated on observed entries.

\section{PRELIMINARIES}

\subsection{Neural Tucker Factorization}

Neural Tucker factorization (NeuTucF) expresses element-wise approximation through Tucker-style latent interactions. For a three-way tensor, it can be written as
\begin{equation}
{\widehat{y}}_{ijk} = \mathop{\sum}\limits_{{p = 1}}^{P}\mathop{\sum}\limits_{{q = 1}}^{Q}\mathop{\sum}\limits_{{r = 1}}^{R}\mathcal{G} \odot{\mathcal{T}}_{ijk} = 
\mathop{\sum}\limits_{{p = 1}}^{P}\mathop{\sum}\limits_{{q = 1}}^{Q}\mathop{\sum}\limits_{{r = 1}}^{R}{g}_{pqr}{t}_{pqr}^{\left ( ijk\right ) },
\label{eq:one}
\end{equation}
where $ \cdot $ and $ \odot $ represent the inner product and Hadamard product operations between two tensors, respectively.$ {\mathcal{T}}_{ijk} $ is a rank one interactive tensor derived from the outer product of three potential eigenvectors $ {a}_{i} \in  {\mathbb{R}}^{P},{b}_{j} \in  {\mathbb{R}}^{Q} $ and $ {c}_{k} \in  {\mathbb{R}}^{R} $

\subsection{Selective State Space Model}

A state space model represents sequential dynamics through a latent state; we follow the formulation used in Mamba. It maps the input x(t) to the output y(t) by hiding the state h(t). The internal state of the system is controlled by a first-order differential equation. The specific process is as follows
\begin{equation}
\begin{aligned}
h(t)' &= \mathbf{A}h(t) + \mathbf{B}x(t), \\
y(t) &= \mathbf{C}h(t),
\end{aligned}
\label{eq:two}
\end{equation}
where $ \mathbf{A} \in  {{R}}^{N \times  N} $ is the state evolution matrix,$ f{B},\mathbf{C} \in  {{R}}^{N \times  D} $ are projection parameters.

In order to process discrete sequence data, continuous parameters A and B can be discretized by zero-order holding and time sampling at intervals of $ \Delta $ as follows
\begin{equation}
\begin{aligned}
\overline{\mathbf{A}} &= e^{\mathbf{\Delta}\mathbf{A}}, \\
\overline{\mathbf{B}} &= \bigl(\mathbf{\Delta}\mathbf{A}\bigr)^{-1}
\bigl(e^{\mathbf{\Delta}\mathbf{A}} - \mathbf{I}\bigr) \cdot \mathbf{\Delta}\mathbf{B}.
\end{aligned}
\label{eq:three}
\end{equation}
then the discrete form of \eqref{eq:two} is expressed as
\begin{equation}
\begin{aligned}
h_t &= \overline{\mathbf{A}} h_{t-1} + \overline{\mathbf{B}} x_t, \\
y_t &= \mathbf{C} h_t.
\end{aligned}
\label{eq:four}
\end{equation}
where $\overline{\mathbf{A}} \in \mathbb{R}^{N_s \times N_s}, \overline{\mathbf{B}} \in \mathbb{R}^{N_s \times D_s}.$
When the parameters are time-invariant, the recurrence in \eqref{eq:four} admits a convolutional representation. Input-dependent selection instead uses a hardware-aware parallel scan.

Conventional state space models  are designed to capture sequential context with linear time complexity, yet they are inherently limited by static parameterization and lack the capacity for content-based reasoning. To address this limitation, Gu and Dao proposed Mamba with input-dependent selective state spaces. It enables the model to selectively propagate or discard information along the sequence conditioned on the current input content, by parameterizing the core SSM parameters as a function of the inputs. Specifically, the key parameters \(B\), \(C\) and \(\Delta\) in S6 are directly derived from the input sequence \(x(t)\), realizing sequence-adaptive parameterization for dynamic sequential modeling.

\section{Method}
As shown in Figure 1, the proposed Mamba based neural Tucker factorization model (S3NTF) is composed of three components: a model encoding layer, a Mamba based time encoder, and an latent spatiotemporal feature interactive learning module. We will describe these modules in detail in the following sections.
\begin{figure}[htbp]
  \centering
  \includegraphics[width=\linewidth]{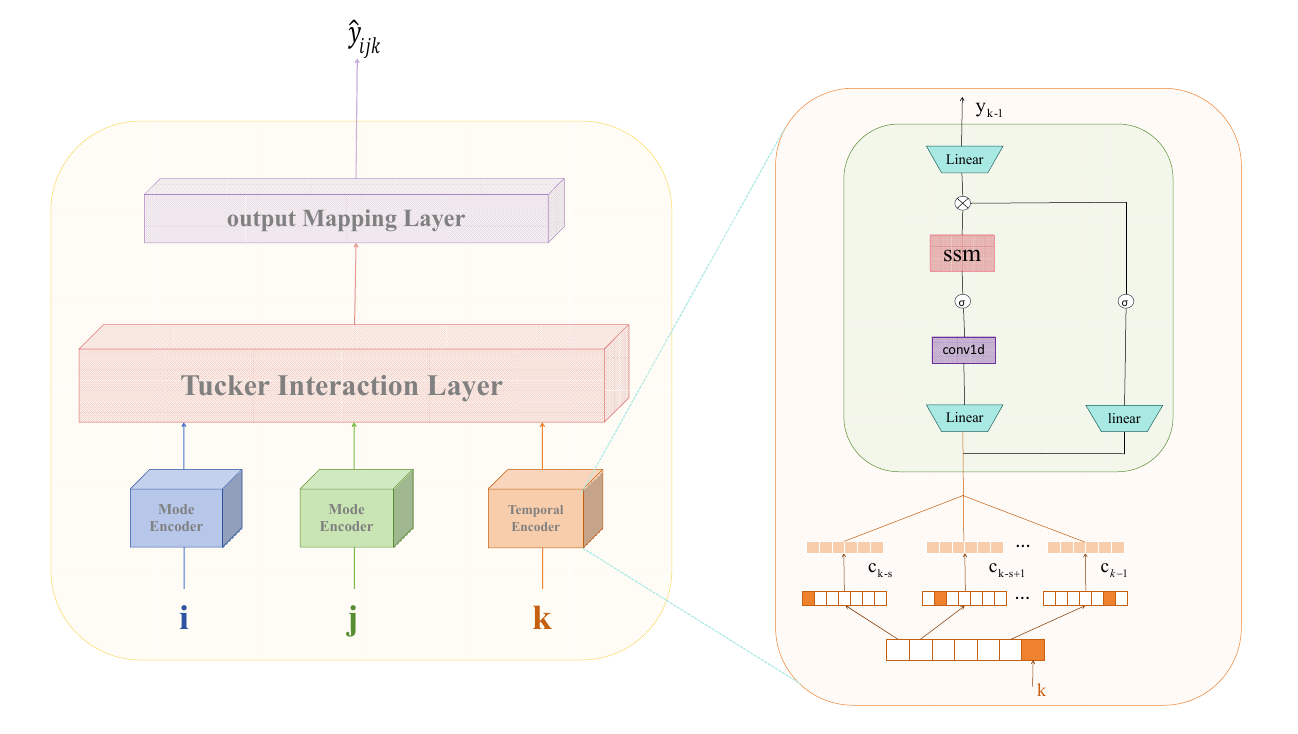}
  \caption{Selective State Space Neural Tucker Factorization Model (S3NFT). The left half describes the neural network structure that simulates complex spatiotemporal interactions, while the right half describes a Mamba based time encoder that captures inherent dynamics along the time dimension.}
  \label{fig:mamba}
\end{figure}

\subsection{Model Encoding Layer} 
Our model is designed for precise spatiotemporal representation learning on high-dimensional interaction (HDI) tensors $ \left( {\mathcal{A} \in  {\mathbb{R}}^{{M}_{1} \times  {M}_{2} \times  \cdots  \times  {M}_{n}}}\right) $. Each pattern corresponds to a temporal or spatial dimension.This article focuses on commonly used three-dimensional tensor structures$ \mathcal{A} \in  {\mathbb{R}}^{I \times  J \times  K} $

In the input phase, we follow the density-oriented index-embedding construction of NeuTucF. Specifically, a single hot coding layer first converts the high cardinality spatiotemporal index into its corresponding sparse vector through the single hot coding process, and then maps the sparse single hot coding of the pattern index to its respective dense set embedding vector through a linear layer

For index mode input $ i \in  {\mathbb{Z}}^{ + } $, the specific process can be expressed as
\begin{equation}
\begin{aligned}
e_i &= \mathcal{F} e(i), \\
a_i &= \mathcal{A} e_i.
\end{aligned}
\label{eq:five}
\end{equation}
they are respectively the process of one-hot encoding and dense embedding.Where $ {\mathcal{F}}_{e}\left( \cdot \right) $ represents the operation that generates a one-hot vector $ {\mathbf{e}}_{i} \in  {\mathbb{R}}^{I} $ for a particular pattern input with index $ i $, and $ {\mathbf{a}}_{i} \in  {\mathbb{R}}^{P} $ represents its respective embedding vector.$ \mathbf{A} \in  {\mathbb{R}}^{P \times  I} $ is the learnable weight matrix used in the dense embedding process, where I denotes the maximum cardinality of the corresponding pattern and P denotes the embedding dimension.

By mapping indexed spatial and temporal coordinates into compact, low-dimensional embeddings, this encoder effectively alleviates the curse of high-dimensional sparsity while capturing underlying spatial and temporal semantics in an implicit manner. The resulting latent representations are lightweight, interpretable, and adaptable to downstream tasks, providing subsequent neural modules with robust inputs that enhance both the fidelity of spatiotemporal pattern learning and the overall computational efficiency.

\subsection{Temporal Encoder}
Following the historical temporal-embedding construction in GARNT, we employ a Mamba-based encoder to process the temporal mode $k$ of the three-dimensional input. Specifically, for a given time step k , we leverage the embedding vectors from the preceding s time steps as input. These embeddings are also generated by Eq.\eqref{eq:five}, forming a time series that captures historical traffic dynamics.

Let ${c}_{t}\in {R}^{R}$ denote the dense embedding vector for temporal index t  obtained from the Eq.\eqref{eq:five}, where R  is the temporal embedding dimension. The input sequence to the Mamba encoder is constructed as
\begin{equation}
\mathcal{S}_{k} = [\mathbf{c}_{k-s}, \mathbf{c}_{k-s+1}, \ldots, \mathbf{c}_{k-1}] \in \mathbb{R}^{s \times R},
\label{eq:input_seq}
\end{equation}
where s represents the historical time window size and ${S}_{k}$ denotes the stacked embedding matrix for time step k .

The Mamba block first maps the input features into an extended hidden space of dimension \textit{ED} through a linear projection layer,obtaining ${x}$ and ${z}$. ${E}$ denotes the expansion factor, ${D}$ denotes the model dimension,here ED=ER.

Subsequently, convolutional layers and silu activation functions are applied to extract local positional correlations to obtain ${x}^{'}$. This process can be represented as
\begin{equation}
    {x}^{'}=SiLU(Conv1D(x))
\end{equation}
where the convolution operation enhances the representation's robustness to local temporal structural details.

The core of the Mamba block lies in a discretized selective state space model. Driven by input parameters, this SSM module selectively encodes long-range sequence patterns by utilizing state transition dynamics and input related gating mechanisms. The specific details of the selective space model are shown in Eq.\eqref{eq:four}. Through this process, we can obtain the output ${y}$.
\begin{equation}
    {y}=SelectiveSSM(\overline{A},\overline{B},c)({x}^{'})
\end{equation}
Following the Mamba block design, the SSM output is gated by the SiLU-activated $z$ branch and then passed through an output projection.

We obtain the final output of the Mamba encoder with a time step k by processing the entire historical sequence. This process can be described as:
\begin{equation}
\hat{\mathbf{T}_{k}} = \mathbf{y}_{k-1} = \text{Mamba}(\mathbf{c}_{k-s}, \mathbf{c}_{k-s+1}, \ldots, \mathbf{c}_{k-1}),
\label{eq:temporal_output}
\end{equation}
where  $\hat{{T}_{k}}$ represents the temporally-aware feature embedding that integrates rich long-term dependencies. This embedding is subsequently fed into the latent spatiotemporal feature interaction learning module, where it interacts with spatial embeddings.

\subsection{Latent Spatiotemporal Feature Interactive Learning Module}

Through the model encoding layer and the time encoder based on mamba, we respectively obtained the dense pattern embeddings $ {\mathbf{a}}_{i} \in  {\mathbb{R}}^{P},{\mathbf{b}}_{j} \in \; {\mathbb{R}}^{Q} $ and $ \hat{{T}_{k}} \in  {\mathbb{R}}^{R} $ for the three input patterns with embedding dimensions of P, Q, and R.

Following the Tucker interaction construction in NeuTucF, we take the outer product of the mode embeddings to form a rank-one interaction tensor.For the three-way case, the rank-one interaction tensor ${\mathcal{T}_{ijk}}$ is formally defined as the outer product  of the i-th, j-th, and k-th  vectors in the form of
\begin{equation}
{\mathcal{T}}_{ijk} = {\mathbf{a}}_{i} \circ{\mathbf{b}}_{j} \circ\hat{{T}_{k}} , 
\label{eq:seven}
\end{equation}
where $ \circ $ represents outer product operator.At the same time, we obtain an outer product interaction tensor $ {\mathcal{T}}_{ijk} \in  {\mathbb{R}}^{P \times  Q \times  R} $.Each element of ${\mathcal{T}_{ijk}}$ represents the interaction between potential factors ${a}_{ip}$, ${b}_{jq}$ and ${c}_{kr}$.

In order to make it compatible with any other neural network module,the interaction tensor is flattened into an interaction vector, which can be expressed as
\begin{equation}
{\mathbf{t}}_{ijk} = \mathcal{F}\left ( {\mathcal{T}}_{ijk}\right ) ,
\label{eq:eight}
\end{equation}
where $\mathcal{F}$ denotes the flatten operation,The output $ {t}_{ijk} \in \; {\mathbb{R}}^{{d}_{\text{ model }}} $ is a tucker interaction vector of size $ model = P \times  Q \times  R $.

Finally, $ {\mathbf{t}}_{ijk} $ is projected into predictive output through linear transformation, which is formalized as
\begin{equation}
{\widehat{y}}_{ijk}= {\phi}_{o}({t}_{ijk})=\sigma\left ( {\mathbf{W}{\mathbf{t}}_{ijk}}\right ) ,
\label{eq:nine}
\end{equation}
where $ {\phi }_{o}\left( \cdot \right) $ is the output function of the output layer, which is composed of an activation function $ \sigma \left( \cdot \right) $ and a simple weight matrix $ \mathbf{W} \in  {\mathbb{R}}^{1 \times  {d}_{\text{ model }}} $ of the linear layer.
The present implementation employs a basic linear projection layer as the final output transformation, primarily to validate the efficacy of integrating Mamba-based temporal modeling for capturing long-range dynamic dependencies within the neural Tucker factorization framework. While this minimalist design successfully demonstrates the core hypothesis of Mamba-enhanced temporal-aware tensor factorization, future investigations could explore substituting the linear layer with advanced architectures such as multilayer perceptrons (MLPs), multi-head self-attention mechanisms. Such extensions would enable a systematic evaluation aimed at uncovering richer spatiotemporal patterns from tensor data, potentially revealing stronger synergies between long-sequence modeling and high-order spatiotemporal representation learning.

\subsection{Learning Scheme}
In order to verify the completion ability of the model on the traffic dataset, the objective function is set to minimize the difference between the original tensor $ \mathcal{Y} $ and the approximate value $ \widehat{\mathcal{Y}} $ using Euclidean distance. Since only information from the known element set $ \Lambda $ needs to be described, the objective function is formally defined as follows:
\begin{equation}
\begin{aligned}
\mathcal{E} &= \frac{1}{2}\sum_{y_{ijk} \in \Lambda} \left( y_{ijk} - \widehat{y}_{ijk} \right)^2, \\
&= \frac{1}{2}\sum_{y_{ijk} \in \Lambda} \left( y_{ijk} - \sigma\left( \mathbf{W} \mathbf{t}_{ijk} \right) \right)^2.
\end{aligned}
\label{eq:ten}
\end{equation}
it is the objective that the model needs to minimize. The model parameters can be trained using stochastic gradient descent (SGD) or its adaptive variants such as AdaGrad, Adam, and RMSProp. This article uses Adam optimizer for optimization.

\begin{table}[htbp]
  \centering
  \caption{Details of Experiment Datasets}
  \label{tab:dataset_details}
  \renewcommand{\arraystretch}{2.5}
  \begin{tabular}{l c c c c}
    \hline
    No. & Dataset & Dimension & Known Count & Density (\%) \\
    \hline
    D1 & \makecell{Seattle City \\ Speed} & 323$\times$28$\times$288 & 260,467 & 9.9998 \\
    \hline
    D2 & \makecell{New York \\ City Flow} & 30$\times$30$\times$1064 & 97,446 & 7.3957 \\
    \hline
    D3 & \makecell{Hangzhou \\ Flow} & 80$\times$105$\times$28 & 20,976 & 8.9183 \\
    \hline
    D4 & \makecell{Guangzhou \\ Speed} & 214$\times$61$\times$144 & 185,559 & 9.8713 \\
    \hline
    D5 & \makecell{METR-LA \\ Speed} & 288$\times$119$\times$207 & 651,900 & 9.1891 \\
    \hline
    D6 & \makecell{PEMS-BAY \\ Speed} & 288$\times$181$\times$325 & 1,693,717 & 9.9973 \\
    \hline
  \end{tabular}
\end{table}

\section{EXPERIMENTS}

\subsection{Experiment Setting}
The experiment was conducted on a platform equipped with a 3.2-GHz AMD Ryzen 7 5800h with Radeon graphics CPU and an NVIDIA GeForce  RTX 3056 laptop GPU.All the model are implemented with Python 3.10.19,Pytorch 2.1.1,causal-conv1d 1.2.1 and mamba-ssm 2.2.2.In this paper, the average absolute error (MAE), the average relative error (MRE) and the root mean square error (RMSE) are used for evaluation.

We conducted experiments on real traffic data sets in six different cities.In all experiments, in order to make the data closer to the normal distribution, we preprocessed the data by logarithmic transformation and min max normalization.

The baseline definitions and reported results for M1--M7 follow the comparison protocol of GARNT. The comparison models are as follows:
\begin{enumerate}
    \item M1: Lightweight Graph Convolutional Network (LightGCN)~\cite{he2020lightgcn};
    \item M2: a Nonnegative Canonical Polyadic-based Latent Factorization of Tensor (NNCP) model~\cite{zhang2014nncp};
    \item M3: a Canonical Polyadic-based Latent Factorization of Tensor model with Cauchy loss-based objective function (CTF)~\cite{ye2021ctf};
    \item M4: Biased Nonnegative Latent Factorization of Tensor (BNLFT) model with SLF-NMUT and ADM-STL Optimization~\cite{r13};
    \item M5: a Neural Collaborative Filtering model with Matrix Factorization and Neural Network Integration (NCF)~\cite{he2017ncf};
    \item M6: a Neural Tucker Factorization model with Linear Layer and Sigmoid Output Projection (NeuTucF)~\cite{r73};
    \item M7: a Gated-Recurrent Neural Tucker Factorization (GARNT)~\cite{hou2025tntf};
    \item M8: Mamba-based Neural Tucker Factorization model(S3NTF).
\end{enumerate}

To ensure a fair and rigorous comparison with the state-of-the-art baseline methods, we strictly follow the experimental setup, hyperparameter configurations, and evaluation protocols reported in the GARNT paper. Specifically, we directly adopt the publicly reported experimental results of all baseline models (including GARNT) on the six real-world traffic datasets, which are obtained under the same data preprocessing, missing rate settings, and evaluation metrics as our proposed S3NTF model. For consistency, we maintain identical embedding sizes, tensor ranks, and training configurations as specified in the original work: the embedding sizes of all neural network-based models are set to 5, and the ranks of the LFT models are uniformly fixed at 5. Additionally, consistent with the original implementation, the LFT models (M2-M4) are augmented with SLF-NMUT and ADM-STL, and their raw outputs are clipped to the [0,1] interval before inverse transformation to ensure value range consistency across all comparative methods. All hyperparameters of our proposed model are tuned via grid search to achieve optimal performance.
\begin{table*}[!t]
  \centering
  \caption{Performance Comparison of Different Models. M1--M7 results are taken from GARNT; M8 denotes the proposed model.}
  \label{tab:model_performance}
  \resizebox{\textwidth}{!}{
  \renewcommand{\arraystretch}{1.3} 
  \begin{tabular}{l *{18}{c} c c}
    \toprule
    \multirow{2}{*}{Model} & \multicolumn{3}{c}{D1} & \multicolumn{3}{c}{D2} & \multicolumn{3}{c}{D3} & \multicolumn{3}{c}{D4} & \multicolumn{3}{c}{D5} & \multicolumn{3}{c}{D6} & \multirow{2}{*}{Win/Loss} & \multirow{2}{*}{Rank} \\
    \cmidrule(lr){2-4} \cmidrule(lr){5-7} \cmidrule(lr){8-10} \cmidrule(lr){11-13} \cmidrule(lr){14-16} \cmidrule(lr){17-19}
    & MAE & MRE & RMSE & MAE & MRE & RMSE & MAE & MRE & RMSE & MAE & MRE & RMSE & MAE & MRE & RMSE & MAE & MRE & RMSE & & \\
    \midrule
    M1 & 7.3406 & 0.2282 & 10.5773 & 11.6017 & 4.2979 & 14.9034 & 70.7900 & 1.9060 & 123.2378 & 6.5716 & 0.2198 & 8.6559 & 8.9371 & 0.2377 & 12.1656 & 5.0974 & 0.1181 & 9.2495 & 0/18 & 8.00 \\
    M2 & 5.4235 & 0.1356 & 7.8008 & 5.0889 & 0.6151 & 11.0201 & 25.2415 & 0.2217 & 59.7254 & 3.6128 & 0.1238 & 5.1250 & 6.0145 & 0.1482 & 8.8401 & 3.9895 & 0.0816 & 6.1690 & 0/18 & 6.17 \\
    M3 & 5.4649 & 0.1371 & 7.8439 & 5.1034 & 0.6310 & 10.9229 & 25.2210 & 0.2210 & 58.9092 & 3.6610 & 0.1257 & 5.1598 & 5.9879 & 0.1476 & 8.8081 & 3.9887 & 0.0821 & 6.1943 & 0/18 & 6.28 \\
    M4 & 4.9711 & 0.1293 & 7.2974 & 5.0559 & 0.6146 & 10.9956 & 24.9841 & 0.2425 & 55.1600 & 3.6676 & 0.1256 & 5.1623 & 5.7189 & 0.1426 & 8.6033 & 3.3521 & 0.0702 & 5.5257 & 0/18 & 5.39 \\
    M5 & 4.5040 & 0.1148 & 6.7490 & 4.2742 & 0.5307 & 8.4263 & 24.9185 & 0.2175 & 55.8657 & 3.2746 & 0.1129 & 4.7575 & 4.7763 & 0.1184 & 7.5979 & 3.1842 & 0.0667 & 5.3216 & 0/18 & 3.83 \\
    M6 & 4.3388 & 0.1073 & 6.5230 & 4.2290 & 0.5062 & 8.4652 & 25.3247 & 0.2200 & 54.8354 & 3.2523 & 0.1125 & 4.7324 & 4.7251 & 0.1177 & 7.5658 & 3.2599 & 0.0663 & 5.2439 & 0/18 & 3.39 \\
    M7 & 4.2152 & 0.1050 & 6.4209 & 4.1703 & 0.4908 & 8.2335 & 24.7456 & 0.2160 & 52.8183 & 3.2364 & 0.1120 & 4.7179 & 4.6832 & 0.1176 & 7.5484 & 3.0995 & 0.0634 & 5.1017 & 0/18 & 2.00 \\
    M8 & \textbf{4.1760} & \textbf{0.1042} & \textbf{6.3774} & \textbf{4.1123} & \textbf{0.4761} & \textbf{8.2015} & \textbf{24.1064} & \textbf{0.2144} & \textbf{52.6424} & \textbf{3.2256} & \textbf{0.1119} & \textbf{4.7130} & \textbf{4.6585} & \textbf{0.1171} & \textbf{7.5146} & \textbf{3.0143} & \textbf{0.0627} & \textbf{5.0894} & \textbf{18/18} & \textbf{1.00} \\
    \bottomrule
  \end{tabular}
  }
\end{table*}

\subsection{Result Analysis}

Table II systematically presents the comprehensive experimental results with statistics-win/loss tallies and Friedman ranking, from which we can see.

The completion performance of all comparison models on six real traffic datasets (D1-D6) is shown in Table II. The M8 model proposed in this article achieved comprehensive optimal performance, achieving leading results on all six datasets, fully verifying the effectiveness of the Mamba selective state space model and NeuTucF fusion, and proving that Mamba's long sequence modeling ability can significantly improve the accuracy of traffic data completion. From the perspective of model categories, the overall performance of deep learning based temporal enhanced neural tensor decomposition models (M5-M8) is far superior to traditional low rank tensor decomposition methods (M2-M4) and basic neural network models (M1). Among them, M8, M7 (GARNT), and M6 (NeuTucF) respectively ranked in the top three, verifying the superiority of the neural Tucker decomposition framework in traffic data completion tasks and confirming the key role of the temporal modeling module in improving completion performance.On the long-term traffic dataset PEMS-BAY (D6), our method reduced the root mean square error (RMSE) by 44.98\%, 17.50\%, 17.84\%, 7.90\%, 4.36\%, 2.95\%, and 0.24\% compared to M1-M7, respectively, fully demonstrating the effectiveness of our model in representing complex spatiotemporal interactions. It may become a new paradigm for learning accurate spatiotemporal representations in traffic analysis.

The hyperparameter configuration on state space dimension and preceding time steps in the Mamba architecture directly governs the fidelity of long-range temporal feature extraction for traffic data. Experimental results reveal distinct dependencies between the key hyperparameters of our Mamba-based temporal encoder and imputation performance in the context of traffic data completion. As shown in Fig. 2a, the model performance exhibits a fluctuating yet overall optimizing trend with the increase of the state space dimension. On the D2 dataset, both RMSE and MAE reach their lowest values when the state space dimension is set to 32, while insufficient dimension settings (ranging from 2 to 16) lead to consistently higher imputation errors. This phenomenon indicates that an under-sized state space will restrict the model’s ability to encode complex long-range temporal dependencies in traffic sequences, while a properly expanded state dimension enables the Mamba block to capture fine-grained sequential patterns more effectively. Fig. 2b shows that increasing the number of preceding time steps causes the prediction accuracy to fluctuate. On D2, both MAE and RMSE achieve their optimal performance at 16 preceding steps. Higher errors under smaller step settings indicate that insufficient temporal information is captured when using too few preceding steps, while error fluctuations under medium step sizes suggest that noise and redundancy may dominate when the step length exceeds the effective temporal correlation range of traffic data. This highlights the importance of balancing the preservation of useful sequence information with the suppression of noise or redundancy.

\begin{figure}[htbp]
  \centering
  \includegraphics[width=0.5\textwidth]{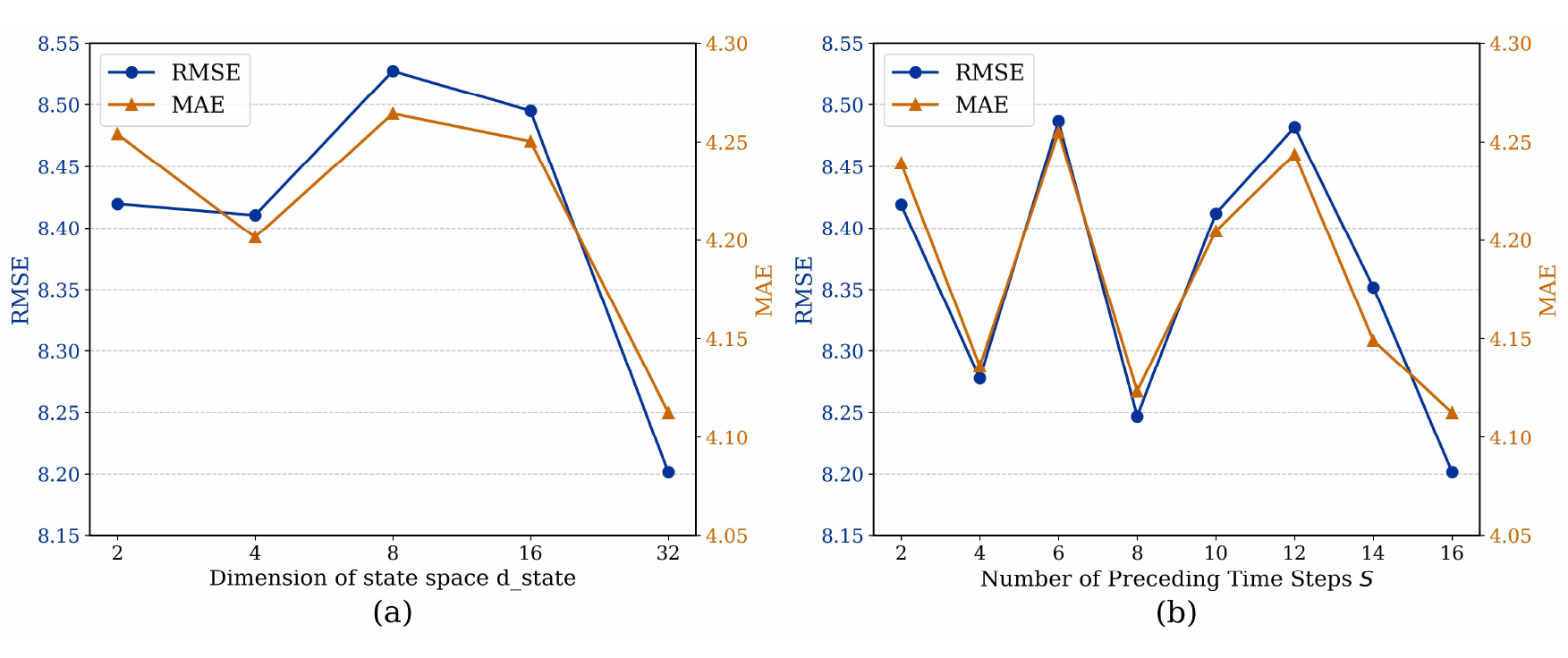}
  \caption{The effect of  state dimension and the number of time step in Mamba on performance of S3NTF on D2.}
  \label{fig:The Influence of State Space Dimension and Time Step on Model Performance.}
\end{figure}

\section{CONCLUSIONS}
This article proposes a selective state space neural Tucker factorization (S3NFT) model. This architecture will integrate Mamba-based temporal encoder with neural Tucker decomposition to dynamically model traffic patterns that evolve over time while preserving interpretable spatial correlations. The Mamba based temporal encoder adaptively captures long-range temporal dependencies through a selective state transition mechanism of input perception. Through a content related information propagation and forgetting mechanism, it effectively models the periodic fluctuations and sudden congestion anomalies of traffic flow, while overcoming the inherent defects of vanishing gradients and poor parallelism in the cyclic architecture.

At the same time, the latent spatiotemporal feature interaction module constructs a factorized implicit representation space through modal exclusive embedding outer product, explicitly characterizing the high-order cross dependencies between spatial segments and temporal dynamics, without losing the inherent interpretability of tensor decomposition. This integrated framework combines the mathematical interpretability and low rank scalability of Tucker decomposition, as well as the superior long sequence modeling ability of modern state space models.

Future work will explore several directions. Firstly, a bidirectional Mamba or layered Mamba architecture can be designed to capture multi-scale temporal dynamics, such as decomposing the patterns of each day and hour of the week into separate encoding paths. Secondly, the linear projection output layer can be replaced with more advanced readout structures, including multi head self attention or autoencoder based modules, to further utilize the rich spatiotemporal representation generated by the Mamba Tucker core. These explorations will drive the development of robust and efficient spatiotemporal learning frameworks, providing reliable data-driven decision support for intelligent traffic control in smart city scenarios.

\bibliographystyle{IEEEtran}
\bibliography{r1,traffic}

\end{document}